\pdfoutput=1

\documentclass[11pt]{article}

\usepackage[review]{acl}

\usepackage{times}
\usepackage{latexsym}
\usepackage{CJK}
\usepackage{multirow}
\usepackage{microtype}
\usepackage{graphicx}
\usepackage{ulem}
\usepackage{amssymb}
\usepackage{booktabs}
\usepackage[T1]{fontenc}

\usepackage[utf8]{inputenc}

\usepackage{microtype}
\usepackage{url}
\title{Improving Pre-trained Language Models with Syntactic Dependency Prediction Task for Chinese Semantic Error Recognition}

\author{Bo Sun\textsuperscript{1}\thanks{\quad indicates equal contribution}, Baoxin Wang\textsuperscript{1,2*}, Wanxiang Che\textsuperscript{1}\thanks{\quad Corresponding Author: W.Che (car@ir.hit.edu.cn)}, Dayong Wu\textsuperscript{2}, Zhigang Chen\textsuperscript{2}, Ting Liu\textsuperscript{1}\\
\textsuperscript{1}Research Center for SCIR, Harbin Institute of Technology, Harbin, China\\
\textsuperscript{2}State Key Laboratory of Cognitive Intelligence, iFLYTEK Research, China\\
\texttt{\{bsun, car, tliu\}@ir.hit.edu.cn}\\
\texttt{\{bxwang2, dywu2, zgchen\}@iflytek.com}
}

\begin{document}
\maketitle
\begin{abstract}
Existing Chinese text error detection mainly focuses on spelling and simple grammatical errors. These errors have been studied extensively and are relatively simple for humans. On the contrary, Chinese semantic errors are understudied and more complex that humans cannot easily recognize. The task of this paper is Chinese Semantic Error Recognition (CSER), a binary classification task to determine whether a sentence contains semantic errors. The current research has no effective method to solve this task. In this paper, we inherit the model structure of BERT and design several syntax-related pre-training tasks so that the model can learn syntactic knowledge. Our pre-training tasks consider both the directionality of the dependency structure and the diversity of the dependency relationship. Due to the lack of a published dataset for CSER, we build a high-quality dataset for CSER for the first time named Corpus of Chinese Linguistic Semantic Acceptability (CoCLSA). The experimental results on the CoCLSA show that our methods outperform universal pre-trained models and syntax-infused models.
\end{abstract}

\section{Introduction}
The recognition of text errors such as Chinese spelling errors \cite{jiang2012rule} and Chinese grammar errors \cite{lee2015overview} is widely mentioned in previous research. However, there is insufficient research on semantic errors in Chinese sentences, including word order, collocation, missing, redundant, confusion, fuzziness, and illogic errors. The recognition of semantic errors has essential applications in education, journalism, and publishing. This paper considers the Chinese Semantic Error Recognition (CSER) task, a binary classification task, to determine whether a Chinese sentence has semantic errors. 

Unlike Chinese Spelling Check (CSC) and Chinese Grammatical Error Diagnosis (CGED), CSER is oriented to more complex incorrect sentence phenomena and requires dependency-based syntactic knowledge for large incorrect sentences. Table \ref{dataset} shows the examples of text errors for different tasks and error types. As shown in Table \ref{dataset}, the error type of the first sentence in CSER task is word order because \begin{CJK}{UTF8}{gbsn}``听取''\end{CJK} (listen) should be placed before \begin{CJK}{UTF8}{gbsn}``讨论''\end{CJK} (discuss) due to the time sequence. The error type of the first sentence in CGED task is also word order. However, this word order error results in inconsistent semantics, and humans can easily detect it. More examples can be seen in Appendix \ref{text_error}, consisting of all types of errors for different tasks. According to Table \ref{dataset}, we can see that Chinese grammatical errors and Chinese spelling errors can be recognized easily by humans. However, the sentences with semantic errors are relatively fluent and even difficult for humans to recognize because these errors usually require the syntactic structure of words to be judged.

\begin{table*}[t]
\centering
\scalebox{0.9}{
\begin{tabular}{llll}
\toprule
Task & Error Type &Sentence & Translation \\ \hline
\multirow{2}{*}{CSC}  & \multirow{2}{*}{Spelling Errors} &\multirow{2}{*}{ \begin{CJK*}{UTF8}{gbsn}个人\sout{触须}(储蓄)卡存款也有利息吗\end{CJK*}}
&  Is there interest on personal \sout{tentacle}\\
& & &(debit) card deposits \\ \hline
\multirow{4}{*}{CGED} & \multirow{2}{*}{Word Order} & \begin{CJK*}{UTF8}{gbsn}从小到大\sout{为了你们}(你们为了)照顾\end{CJK*} 
& Since childhoood, you have paid a lot\\ 
& &\begin{CJK*}{UTF8}{gbsn}我，付出很多\end{CJK*} 
&to take care of me \\\cline{2-4}
&\multirow{2}{*}{Redundant}&\begin{CJK*}{UTF8}{gbsn}也有些人不喜欢流行歌曲\sout{的}，甚至\end{CJK*}&Some people don't like pop songs, and\\ 
& &\begin{CJK*}{UTF8}{gbsn}有些人特别蔑视流行歌曲\end{CJK*}
&some even despise pop songs\\\hline
\multirow{3}{*}{CSER} & \multirow{2}{*}{Word Order} & \begin{CJK*}{UTF8}{gbsn}全厂职工\sout{讨论并听取}(听取并讨论)\end{CJK*} & The whole staff \sout{discuss and listen} \\
& & \begin{CJK*}{UTF8}{gbsn}了报告\end{CJK*}&(listen and discuss) the report\\\cline{2-4}
& \multirow{2}{*}{Collocation} & \multirow{2}{*}{ \begin{CJK*}{UTF8}{gbsn}国内彩电\sout{市场}严重滞销\end{CJK*}}&Domestic TV \sout{market} (sets) are seriously\\
&&& unsalable \\
\bottomrule
\end{tabular}}
\caption{Examples of different tasks. We cannot translate sentences with incorrect semantic in the CGED task, so we translate them into the correct semantic.} 
\label{dataset}
\end{table*}

\begin{figure}[t]
\centering
\includegraphics[width=0.9\columnwidth]{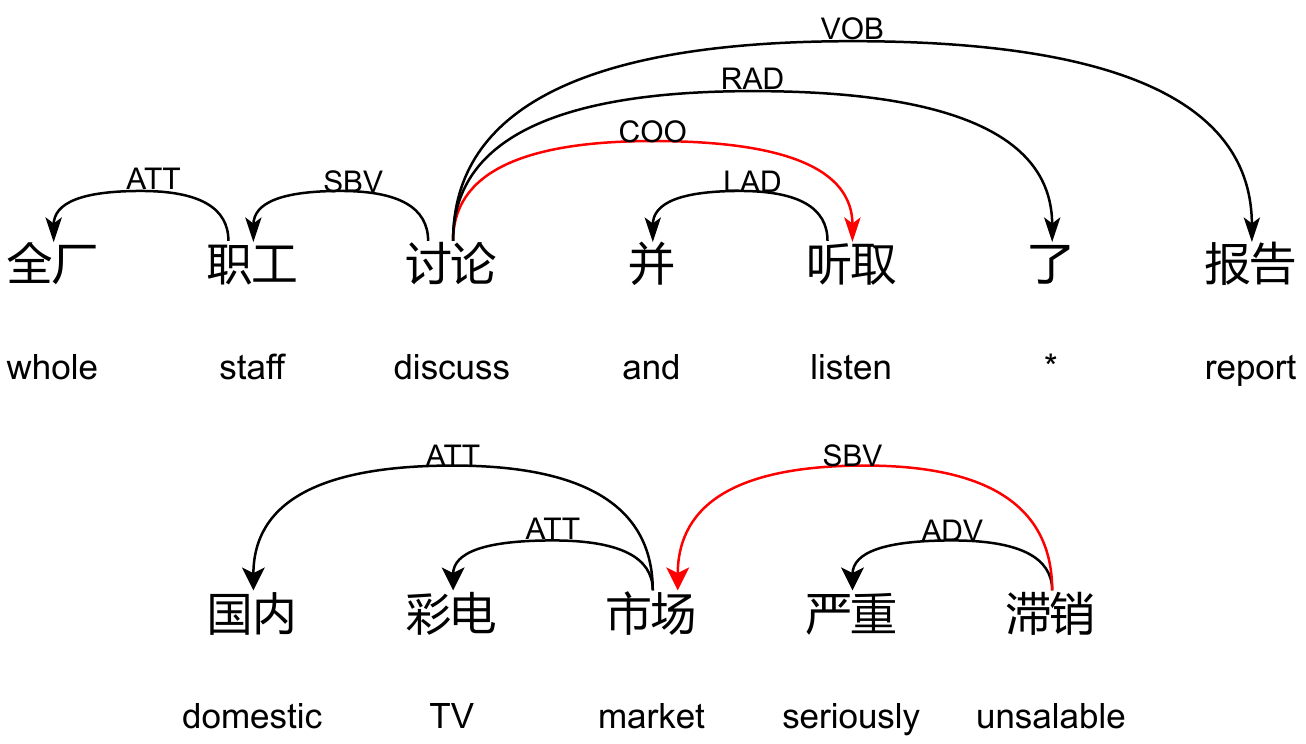} 
\caption{Syntax analysis of incorrect semantic sentences, the incorrect semantic is marked as red.}
\label{fig-eg}
\end{figure}

CSER task benefits from dependency-based syntactic knowledge. For example, as shown in Figure \ref{fig-eg}, the mistake of the sentence is that there is the wrong dependency structure between \begin{CJK}{UTF8}{gbsn}``听取''\end{CJK} (listen) and \begin{CJK}{UTF8}{gbsn}``讨论''\end{CJK} (discuss), which can be discovered by syntactic parsing. In the incorrect sentence, \begin{CJK}{UTF8}{gbsn}``讨论''\end{CJK} (discuss) is the parent node of \begin{CJK}{UTF8}{gbsn}``听取''\end{CJK} (listen). However, in the correct sentence, \begin{CJK}{UTF8}{gbsn}``听取''\end{CJK} (listen) should be the parent node of \begin{CJK}{UTF8}{gbsn}``讨论''\end{CJK} (discuss).
From another perspective, the bottom sentence is incorrect due to the wrong dependency relationship between \begin{CJK}{UTF8}{gbsn}``市场''\end{CJK} (market) and \begin{CJK}{UTF8}{gbsn}``滞销''\end{CJK} (unsalable). In the incorrect sentence, the dependency relationship between \begin{CJK}{UTF8}{gbsn}``市场''\end{CJK} (market) and \begin{CJK}{UTF8}{gbsn}``滞销''\end{CJK} (unsalable) is ``subject-verb'' resulting in an improper collocation. Therefore, it is beneficial for the model to learn the syntactic information in both dependency structure and dependency relationship.

This paper introduces a series of novel pre-training tasks, allowing the model to learn more syntactic knowledge based on dependency structure and dependency relationship. We randomly select a couple of words in a sentence and then predict the dependency of the selected words through the representation of the hidden layer of BERT. Then we perform pre-training on a corpus of one million sentences from Wikipedia. We fine-tune the obtained pre-trained models on a high-quality dataset for the CSER task.

To obtain a high-quality dataset for the CSER task, we use the web crawler to obtain Chinese multiple-choice questions related to incorrect semantic sentences from the online resources of the high school examinations. We extract sentences from the above question bank. Then we organize these data into a dataset with correct sentences and incorrect semantic sentences, named Corpus of Chinese Linguistic Semantic Acceptability (CoCLSA). We fine-tune the pre-trained models on CoCLSA. For the CSER task, the experimental results show that our method with syntax-related pre-training tasks exceeds universal pre-trained models for the CSER task. Experiments also show that our method outperforms models that explicitly incorporate syntactic information.

Our main contributions are summarized as follows:

\begin{itemize}
\item We propose several innovative pre-training tasks to learn the directionality of the dependency structure and the diversity of the dependency relationship without adding additional knowledge explicitly. Experiments have proved that our pre-training tasks perform better on CSER task.

\item We provide a high-quality dataset for CSER with 49,408 sentences, namely CoCLSA, which solves the problem of the vacancy of the dataset for the CSER task.
\end{itemize}

To facilitate this research, the CoCLSA datasets in this paper will be released at \url{https://github.com/HIT-SCIR/CoCLSA}.

\section{Related Work}
\subsection{Text Error Detection}
Many researchers have made outstanding achievements on CSC \cite{zhang2020spelling,wang2021dynamic} and CGED \cite{fu2018chinese}. Existing CSC and CGED models cannot achieve good results for CSER because semantic errors are often difficult compared to other errors. The existing CGED models do not consider the characteristics of semantic errors related to syntactic dependency. Some researchers try to solve CSER based on rules \cite{wu2015research} and the Semantic Knowledge-base \cite{guan2012study,zhang2021research}. However, the traditional method is powerless for some more complex and obscure semantic errors. As far as we know, there are currently almost no investigators researching Chinese Semantic Error Recognition (CSER) through the pre-trained model.

\subsection{Syntax-Infused Models}
Some researchers explicitly incorporate syntactic information into the model structure by changing the attention mechanism. \citet{bai2021syntax} propose Syntax-BERT, which changes the flow of information in a standard BERT network via a syntax-aware self-attention mechanism. \citet{li2020improving} propose syntax-aware local attention (SLA), where the attention scopes are restrained based on the distances in the syntactic structure. These Syntax-Infused models complicate the model structure and make model training slow because every fine-tuning of downstream tasks requires syntactic parsing. In addition, the syntactic information of sentences with semantic errors is an interference term for the model. Our approach emphasizes dependency syntax only in the pre-training stage, without complicating the model and adding syntactic information that interferes with the model. Moreover, our method can further improve these Syntax-Infused models by adding syntactic-related pre-training tasks to the pre-trained models. 

\subsection{Pre-Trained Models}
A lot of pre-training tasks are proved to be effective, such as pre-training tasks in BERT \cite{devlin2018bert}, ERNIE \cite{sun2019ernie}, BERT-wwm \cite{cui2019pre} and RoBERTa \cite{liu2019roberta}. However, the pre-training tasks in these universal pre-trained models do not consider syntactic dependencies. \citet{xu2020syntax} propose SEPREM utilizing the syntax of text in both pre-training and fine-tuning stages. As mentioned above, using the syntax of the wrong text in the fine-tuning stage is noise to the model. \citet{wang2020k} put forward K-adapter injecting linguistic knowledge by predicting the head index of each token in the given sentence. However, \citet{wang2020k} only consider the head index of dependencies and does not consider other relationships, such as sibling index, etc. In addition, \citet{wang2020k} does not consider the diversity of dependencies because the dependencies between different tokens are not the same. In this paper, we consider both the directionality and diversity of dependencies.

\section{Methodology}
In this section, we first introduce the different classifications of dependency relations. Then we propose pre-training tasks based on Dependency Prediction (DP) according to the above classifications.

\subsection{Syntax Tree}
\begin{table}[t]
\centering
\scalebox{0.9}{
\begin{tabular}{ll|ll}
\toprule
Tag & Description & Tag & Description \\ \hline
SBV & subject-verb    & FOB&fronting-object \\
ADV & adverbial       &CMP& complement \\
VOB & verb-object     & DBL & double  \\
IOB & indirect-object &ATT & attribute   \\
POB & preposition-object & LAD & left adjunct  \\
COO& coordinate     &RAD&right adjunct \\
\bottomrule
\end{tabular}
}
\caption{Tags and their descriptions for LTP.} 
\label{tags}
\end{table}

\begin{figure*}[t]
\centering
\includegraphics[width=1.9\columnwidth]{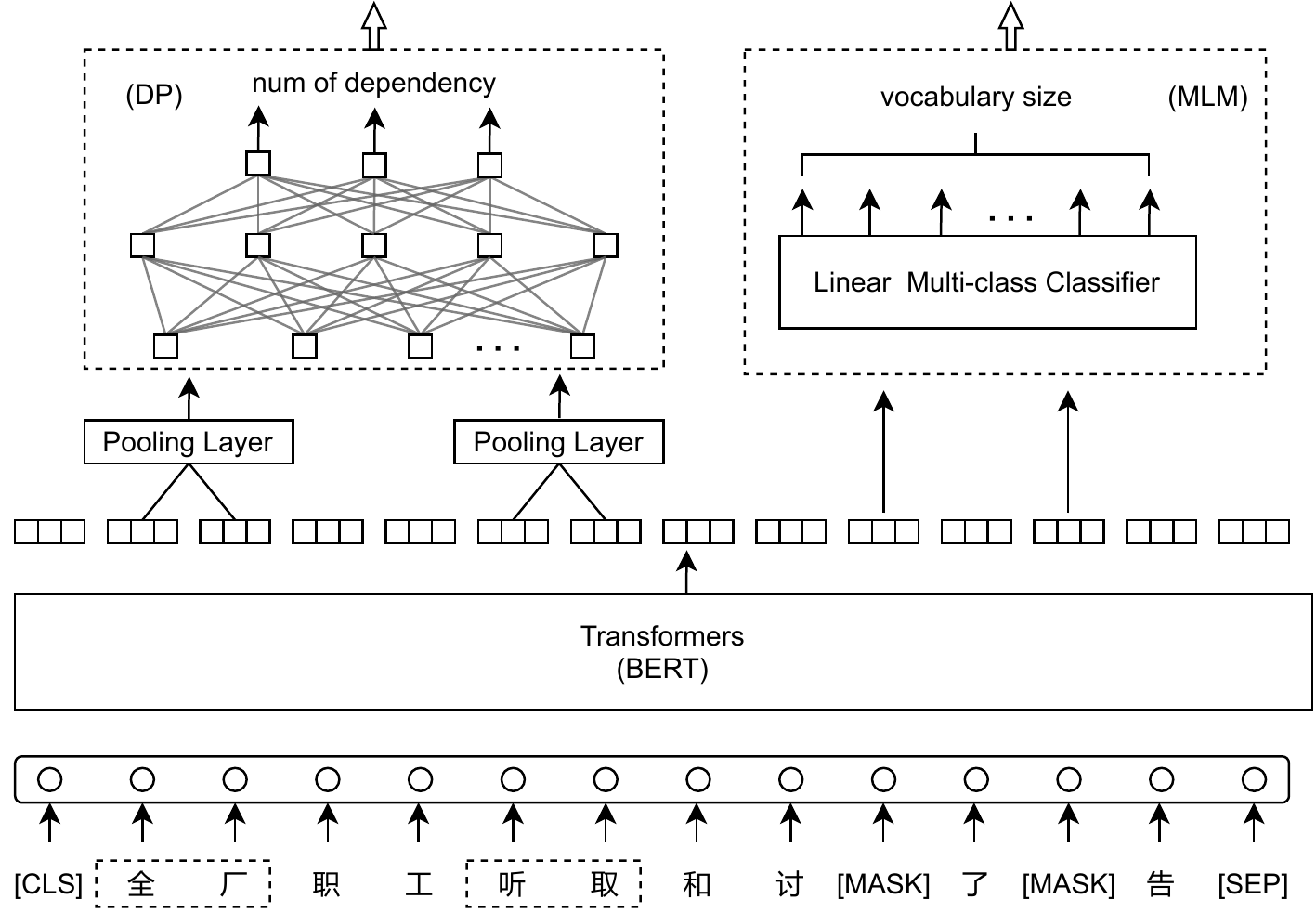} 
\caption{Structure of our pre-trained model.}
\label{pre-task}
\end{figure*}

Dependency parsing shows a substantial improvement in the field of NLP. In this paper, we use the dependency parser of LTP \cite{che2010ltp} to generate dependency parsing, which provides a series of Chinese natural language processing tools. Then we generate the syntax tree based on the dependency parsing results. 
We define the syntax tree as  $\mathcal{T}=\{\mathcal{R},\mathcal{N},\mathcal{E}\}$, where $\mathcal{R}$ represents the relationship between two nodes, $\mathcal{N},\mathcal{E}$ represents node and edge set. We have the following methods to divide classifications of syntactic dependencies:

(1) We consider the directionality of syntactic relationship which is divided into two categories, namely child and parent. The following $\mathcal{D}(\mathcal{N}_i,\mathcal{N}_j)$ is denoted as the length between node $\mathcal{N}_i$ and $\mathcal{N}_j$, which is the minimal length from node $\mathcal{N}_i$ along the edge to node $\mathcal{N}_j$. There are 2 dependency relationships as follows:

\noindent$\mathcal{R}_{ij}=child$ if $\mathcal{N}_i$ is child node of $\mathcal{N}_j$ and $\mathcal{D}(\mathcal{N}_i,\mathcal{N}_j)=1$.

\noindent$\mathcal{R}_{ij}=parent$ if $\mathcal{N}_i$ is parent node of $\mathcal{N}_j$ and $\mathcal{D}(\mathcal{N}_i,\mathcal{N}_j)=1$.

As shown in Figure \ref{fig-eg}, for example, $\mathcal{R}$(\begin{CJK}{UTF8}{gbsn}全厂\end{CJK},\begin{CJK}{UTF8}{gbsn}职工\end{CJK})$=parent$ and $\mathcal{R}$(\begin{CJK}{UTF8}{gbsn}职工\end{CJK},\begin{CJK}{UTF8}{gbsn}全厂\end{CJK})$=child$.

(2) We extend (1) to three types of syntactic relationships, namely child, parent, and others. These relationships can cover all relations in syntactic structure. There are 3 dependency relationships as follows:

\noindent$\mathcal{R}_{ij}=child$ if $\mathcal{N}_i$ is child node of $\mathcal{N}_j$ and $\mathcal{D}(\mathcal{N}_i,\mathcal{N}_j)=1$.

\noindent$\mathcal{R}_{ij}=parent$ if $\mathcal{N}_i$ is parent node of $\mathcal{N}_j$ and $\mathcal{D}(\mathcal{N}_i,\mathcal{N}_j)=1$.

\noindent$\mathcal{R}_{ij}=others$ if any of the above is not met.

As shown in Figure \ref{fig-eg}, for example, $\mathcal{R}$(\begin{CJK}{UTF8}{gbsn}全厂\end{CJK},\begin{CJK}{UTF8}{gbsn}职工\end{CJK})$=parent$, $\mathcal{R}$(\begin{CJK}{UTF8}{gbsn}职工\end{CJK},\begin{CJK}{UTF8}{gbsn}全厂\end{CJK})$=child$  and $\mathcal{R}$(\begin{CJK}{UTF8}{gbsn}了\end{CJK},\begin{CJK}{UTF8}{gbsn}报告\end{CJK})$=others$.

(3) We consider the diversity of syntactic relationship. Through  syntactic dependency parsing, we can find that different words have different dependencies. As shown in Figure \ref{fig-eg}, for example, $\mathcal{R}$(\begin{CJK}{UTF8}{gbsn}职工\end{CJK},\begin{CJK}{UTF8}{gbsn}全厂\end{CJK})$=ATT$ and $\mathcal{R}$(\begin{CJK}{UTF8}{gbsn}讨论\end{CJK},\begin{CJK}{UTF8}{gbsn}职工\end{CJK})$=SBV$. According to the results of dependency  parsing for LTP, we summarize 12 dependency relationships as shown in the Table \ref{tags}.

\subsection{Syntactic Dependency Prediction}

We have the following pre-training tasks as shown in Figure \ref{pre-task}. The first one is MLM, the same as BERT, which can learn the semantic relationship of context. Another pre-training task is Dependency Prediction (DP), which is proposed to allow the pre-trained model to explicitly learn the syntactic information from dependency parsing. We randomly select some pairs of Chinese words and let the model predict the dependency relationship between them. We use BERT to generate the representation of the last hidden states of the pairs of Chinese words we selected. Since Chinese words consist of multiple tokens, we put these pairs of Chinese words into a pooling layer with max-pooling. Then we put it into the classifier for classification tasks. In this paper, we select Multilayer Perceptron (MLP) as the classifier consisting of 4 layers. We select Rectified Linear Unit as an activation function in MLP. We show the structure of our pre-trained model in Figure \ref{pre-task}.

According to the description in Section 3.1, there are three cases for classifying dependencies. Therefore, we have the following pre-training tasks: Dependency Structure Prediction (DSP), Dependency Relation Prediction (DRP), and Dependency Prediction (DP).

\noindent\textbf{DSP}: In this pre-training task, dependency relationship is divided into 2 types, namely $child$ and $parent$. We randomly select some pairs of Chinese words with two dependency relationships using the dependency parser of LTP. These relationships are either $parent$ or $child$. Then we let the model predict these relationships. The pre-trained models can learn the directionality of the dependency relationship in this pre-training task. 

\noindent\textbf{DSP$^{\dagger}$}: Considering that the relationship between two Chinese words is not only $parent$ and $child$, we add a dependency relationship, namely $others$. This dependency relation refers to all relations except $parent$ and $child$. Hence, this pre-training task is an enhanced version of DSP. In this pre-training task, we divide the dependency relationship into 3 types.

\noindent\textbf{DRP}: We can also classify the dependency relationship by labels of syntactic dependencies. In this pre-training task, the dependency relationship is divided into 12 types, as shown in Tabel \ref{tags}. We randomly select some pairs of Chinese words with 12 dependency relationships using the dependency parser of LTP. The pre-trained models can learn the diversity of dependency relationships in this pre-training task.

\noindent\textbf{DP}: We combine DSP and DRP for multi-task training. 

\noindent\textbf{DP$^{\dagger}$}: We combine DSP$^{\dagger}$ and DRP for multi-task training.

\section{Experiments}

\subsection{CoCLSA}
We use the web crawler to obtain Chinese multiple-choice questions related to incorrect semantic sentences from the high school examination online resources. Then we organize these data into a dataset with 49,408 sentences with two labels. One of the labels is correct sentences, and the other is incorrect semantic sentences. We choose 45,248 sentences as the train dataset, 2,160 sentences as the validation dataset, and 2,000 as the test dataset. Since most of the multiple-choice questions we crawl are sentences with semantic errors, there are more incorrect semantic sentences in CoCLSA. We divide the validation and test sets with the same number of correct and incorrect semantic sentences to ensure reasonableness. Therefore, the proportion of incorrect semantic sentences is higher in the training set. We try to use oversampling and undersampling methods to make the number roughly the same in the train set. However, we find that the effect is the same (or worse) as formal training. Hence, we give up using oversampling and undersampling methods.  

In order to prevent the problem of data overlap, we clean the training set: we delete the data whose text similarity between the validation/test sets and the training set is greater than a fixed threshold $\gamma$. We calculate text similarity by Levenshtein Ratio based on Levenshtein Distance. We select the fixed threshold $\gamma=70\%$ because training data whose text similarity is lower than 70\% is of less similarity compared with the validation and test set. As shown in Appendix \ref{overlap}, we enumerate some training data and test data whose similarity is the top-5. We find that the similarity between training data and test data is acceptable, and even some training and test data labels are different.

\begin{table}[t]
\centering
\scalebox{1.0}{
\begin{tabular}{c|ccc}
\toprule
Model   & \#Line & Avg.Length& Error Ratio  \\ \hline
Train   & 45,248 & 50.4 & 74.6\% \\\hline
Dev     & 2,160  & 52.6 & 50.0\% \\\hline
Test    & 2,000  & 54.5 & 50.0\% \\\bottomrule
\end{tabular}}
\caption{Details of CoCLSA where Error Ratio means the proportion of semantic incorrect sentences in the total data.} 
\label{CoCLSA}
\end{table}

\subsection{Experimental setup}
We use 1 million Wikipedia data as a pre-training dataset in the pre-training stage. We use LTP\footnote{\url{http://ltp.ai/}} as a tool for syntactic parsing. We pre-train for 10 epochs with an effective batch size of 256. We use AdamW optimizer \cite{kingma2014adam,loshchilov2017decoupled} with a learning rate of 2e-5. We use a learning rate warmup for 2,500 steps. In the fine-tuning stage, we use CoCLSA as a fine-tuning dataset. We fine-tune the pre-trained models for 4 epochs with an effective batch size of 32. We use AdamW optimizer with a learning rate of 2e-5 and weight decay of 0.01. The implementation of pre-training and fine-tuning is based on HuggingFace's Transformer \cite{wolf2019huggingface}, which consists of 12-layer, 768-hidden, 12-heads. 

\begin{table*}[t]
\renewcommand\arraystretch{1.1}
\centering
\scalebox{1.0}{
\begin{tabular}{lcccc}
\toprule
Model   & $P$& $R$& $F_1$ &$ACC$ \\ \hline
\multicolumn{5}{c}{\textit{Universal Pre-trained Models}}\\\hline
RoBERTa\cite{liu2019roberta} &72.9$\pm$0.5	&72.4$\pm$1.6&72.6$\pm$0.7	&72.7$\pm$0.4\\
MacBERT\cite{cui2020revisiting} &73.3$\pm$0.7	&72.6$\pm$1.9&72.9$\pm$0.6	&73.1$\pm$0.2\\\hline
\multicolumn{5}{c}{\textit{RoBERTa Fine-tuning with Syntax-Infused Models}}\\\hline
SLA\cite{li2020improving}& 72.8$\pm$0.6&73.0$\pm$1.3&72.9$\pm$0.6	&72.9$\pm$0.4\\
Syntax-RoBERTa\cite{bai2021syntax}& 73.3$\pm$0.2&74.3$\pm$0.4&73.8$\pm$0.2	&73.6$\pm$0.1\\ \hline
\multicolumn{5}{c}{\textit{Pre-training with Our Methods}}\\\hline
RoBERTa+DP & \textbf{74.2}$\pm$0.5	&74.4$\pm$1.5&74.3$\pm$0.5	&74.3$\pm$0.2\\
RoBERTa+DP$^{\dagger}$ & 73.2$\pm$1.0	&75.8$\pm$2.1&\textbf{74.8}$\pm$0.3	&74.1$\pm$0.1\\

SLA $+$ DP& 72.1$\pm$1.1&77.1$\pm$1.7&74.5$\pm$0.2	&73.6$\pm$0.3\\
SLA $+$ DP$^{\dagger}$& 72.0$\pm$0.6&76.9$\pm$0.9&74.4$\pm$0.3	&73.5$\pm$0.3\\

Syntax-RoBERTa $+$ DP& 73.7$\pm$0.6	&75.9$\pm$1.3&\textbf{74.8}$\pm$0.4	&\textbf{74.4}$\pm$0.2\\
Syntax-RoBERTa $+$ DP$^{\dagger}$& 73.6$\pm$0.8&\textbf{76.1}$\pm$1.8 &\textbf{74.8}$\pm$0.6	&\textbf{74.4}$\pm$0.3\\\bottomrule
\end{tabular}}
\caption{We report the average score and standard deviation of 3 independent runs with different seeds.}\label{main_results}
\end{table*}

\subsection{Results}
Table \ref{main_results} demonstrates the results of different models on the CSER task fine-tuned with CoCLSA. RoBERTa+DP achieves an improvement of 1.6\% in accuracy score and 1.7\% in F1 score compared with RoBERTa. RoBERTa+DP$^{\dagger}$ achieves an improvement of 1.4\% in accuracy score and 2.2\% in F1 score compared with RoBERTa. The result demonstrates that our DP pre-training task improves the CSER task. 

RoBERTa+DP brings gains of 1.2\% in accuracy score and 1.4\% in F1 score compared with MacBERT. RoBERTa+DP$^{\dagger}$ brings gains of 1.0\% in accuracy score and 1.9\% in F1 score compared with MacBERT. This improvement indicates that our methods outperform general-purpose pre-trained models for tasks that rely heavily on syntactic knowledge, such as CSER.

Comparing to Syntax-RoBERTa, Syntax-RoBERTa+DP (DP$^{\dagger}$) brings an improvement of 0.9\% (0.8\%) in accuracy score and 1.4\% (1.0\%) in F1 score. Comparing to SLA, SLA+DP (DP$^{\dagger}$) brings an improvement of 0.7\% (0.6\%) in accuracy score and 1.6\% (1.5\%) in F1 score. The method in Syntax-Infused models and our method based on novel pre-training tasks are two completely different ideas. Syntax-Infused models explicitly incorporate syntactic information into the model in the fine-tuning stage. On the contrary, we design some dependency-related pre-training tasks to let the model learn syntactic information implicitly in the pre-training stage. This result demonstrates that our method improves Syntax-Infused models by taking our method in the pre-training stage. 

It can also be seen that Syntax-RoBERTa+DP (DP$^{\dagger}$) has less improvement compared to RoBERTa+DP (DP$^{\dagger}$). This result may be partially attributed to incorrect dependency information that can interfere with the model.

\subsection{Ablation Study}
To investigate the impacts of pre-training tasks in RoBERTa+DP and RoBERTa+DP$^{\dagger}$, we conduct experiments as shown in Table \ref{Ablation}. It is worth noting that RoBERTa+DP (DP$^{\dagger}$) can benefit from applying pre-training task DSP (DSP$^{\dagger}$) and DRP for the CSER task. Moreover, RoBERTa+DP without DRP and RoBERTa+DP$^{\dagger}$ without DRP have a similar performance in the CSER task. This result means that pre-training task DSP and DSP$^{\dagger}$ are of similar improvement for DP-BERT. In other words, increasing the type of dependency structure does not improve the model much. This result may be partially attributed to more wrong sites involving the $parent-child$ relationship in DRP for most of the sentences with Chinese semantic errors. Moreover, $others$ relationship in DRP$^{\dagger}$ is more complex compared with $parent-child$ relationship. It is difficult for pre-trained models to learn some valuable information from predicting $others$ relationships.

\begin{table}[t]
\centering
\scalebox{0.8}{
\begin{tabular}{c|cccc}
\toprule
Model   & $P$& $R$& $F_1$ &$ACC$ \\ \hline
RoBERTa &72.9$\pm$0.5	&72.4$\pm$1.6&72.6$\pm$0.7	&72.7$\pm$0.4\\
\hline
\textbf{+DP(ours)} & \textbf{74.2}$\pm$0.5	&74.4$\pm$1.5&74.3$\pm$0.5	&\textbf{74.3}$\pm$0.2\\
w/o DSP & 73.8$\pm$0.6	&73.2$\pm$0.3&73.5$\pm$0.2	&73.7$\pm$0.3\\
w/o DRP & 73.3$\pm$0.3	&74.6$\pm$1.4&73.9$\pm$0.9	&73.7$\pm$0.7\\
\hline
\textbf{+DP$^{\dagger}$(ours)} & 73.2$\pm$1.0	&\textbf{75.8}$\pm$2.1&\textbf{74.8}$\pm$0.3	&74.1$\pm$0.1\\
w/o DSP$^{\dagger}$ & 73.8$\pm$0.6	&73.2$\pm$0.3&73.5$\pm$0.2	&73.7$\pm$0.3\\
w/o DRP & 73.1$\pm$1.1	&75.0$\pm$0.7&74.0$\pm$0.4	&73.7$\pm$0.6\\\bottomrule
\end{tabular}}
\caption{Ablation Study: We report the average score and standard deviation of 3 independent runs with different seeds.} \label{Ablation}
\end{table}

\subsection{Analysis}
\textbf{Does the model also improve on the task of Chinese Grammatical Error Recognition?} To investigate the ability of our model on the task of Chinese Grammatical Error Recognition(CGER), we construct a dataset of CGED as shown in Table \ref{cged}. Unlike CGED, CGER is a binary classification task that focuses on whether a sentence contains grammatical errors, including Redundant, Missing, Selection, and Word Order. We sample an equal number of correct sentences and sentences with grammatical errors from the training sets of CGED 2016 and CGED 2017. Then we deduplicate this part of the data and divide it into a new training set, validation set, and test set. More details can be seen in Table \ref{cged}. 

We can see the experimental results in Table \ref{cged:result}. RoBERTa+DP has an improvement of 1.7\% in accuracy score and 1.8\% in F1 score compared with RoBERTa. RoBERTa+DP$^{\dagger}$ has an improvement of 1.9\% in accuracy score and 2.1\% in F1 score compared with RoBERTa. The result demonstrates that our method also performs well on the CGER task. This improvement is due to the similarity of some types of grammatical and semantic errors. For example, \textit{Word Order} problems not only appear on the CGER task but also on the CSER task. The difference is that \textit{Word Order} errors in CSER are more complex than in CGER and are challenging to identify even by humans.

\begin{table}[t]
\begin{minipage}[t]{0.5\textwidth}
\makeatletter\def\@captype{table}
\centering
\begin{tabular}{cccc}
    \toprule
    Model     & \#Line  & Avg.Length & Error Ratio \\
    \midrule
    Train & 17,623  & 47.0 & 50.1\%  \\
    Dev     & 991   & 47.4 & 50.0\%   \\
    Test    & 991   & 47.4 & 49.7\% \\
    \bottomrule
\end{tabular}
\caption{CGED Dataset}
\label{cged}
\end{minipage}
\begin{minipage}[t]{0.5\textwidth}
\makeatletter\def\@captype{table}
\centering
\begin{tabular}{ccccc}
    \toprule
    Model    & $P$& $R$& $F_1$ &$ACC$ \\
    \midrule
    RoBERTa &  74.0  & 71.6 & 70.9 & 71.6    \\
    RoBERTa+DP & \textbf{75.4} & 73.3 & 72.7 & 73.3      \\
    RoBERTa+DP$^{\dagger}$ & 75.3 & \textbf{73.5} & \textbf{73.0} & \textbf{73.5}  \\
    \bottomrule
\end{tabular}
\caption{Experimental results of our model and baseline for CGED Dataset.}
\label{cged:result}
\end{minipage}
\end{table}

\begin{figure*}[t]
\centering
\includegraphics[width=2.1\columnwidth]{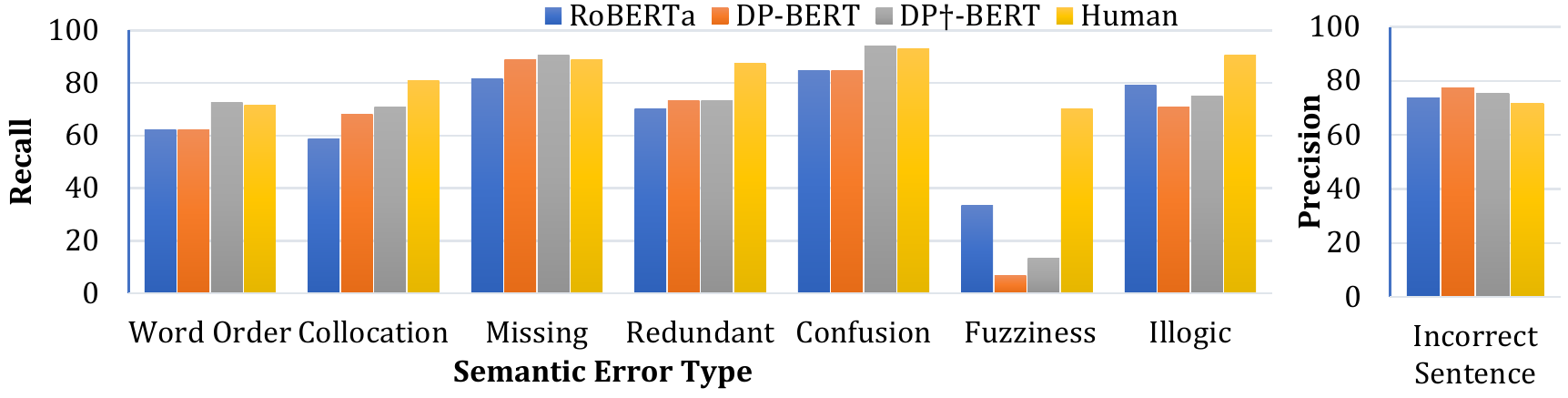} 
\caption{The discriminative ability of the models for different types of semantic errors. We report the average score of 3 independent runs with different seeds for models and the average score of 4 people for human level.}
\label{histogram}
\end{figure*}

\noindent\textbf{How well does the model discriminate under various types of semantic errors?} To figure this out, we randomly sample 200 sentences from our test set, including 100 correct and 100 incorrect sentences. At the same time, to explore college students' ability for CSER task, we hire four students from a top-ranking university and paid remuneration, including two undergraduate students, one master's student, and one doctoral student. In order to ensure the quality of the labeling results, we let these students label the data independently without outside help. We list the precision, recall, F1, and accuracy scores for all types of semantic errors in Appendix \ref{text_error}. Because CSER is a binary classification task, we can only calculate the standard recall score for a specific type of semantic error, as shown in Figure \ref{histogram}.

First, we can see that humans show higher recall scores than almost any model. Compared to our baseline: RoBERTa, our methods exhibit high recall for semantic error types, such as Word Order, Collocation, Missing, Redundant, and Confusion. These error types are strongly related to syntactic information. This result proves that our model does learn practical syntactic knowledge during the pre-training stage. However, our method's recall ability is not as good as the baseline on the semantic error types of Fuzziness and Illogic. These errors have little to do with the syntax but more global semantic information. That is to say, letting the model learn syntactic information cannot solve this kind of problem. That is to say, letting the model learn syntactic information cannot solve this kind of problem but reduces the recall ability of this type of error because the pre-training task focuses on syntax. 

Word Order and Fuzziness are complicated because even humans get lower recall scores of all semantic error types. This may be because people tend to pay less attention to Word Order when speaking in daily life. Some inversions of word order do not affect human understanding of the meaning of sentences, so humans are not so ``strict" on word order issues. Furthermore, Fuzziness is relatively obscure to humans, and these sentences often appear complete.
\begin{figure*}[t]
\centering
\includegraphics[width=2.0\columnwidth]{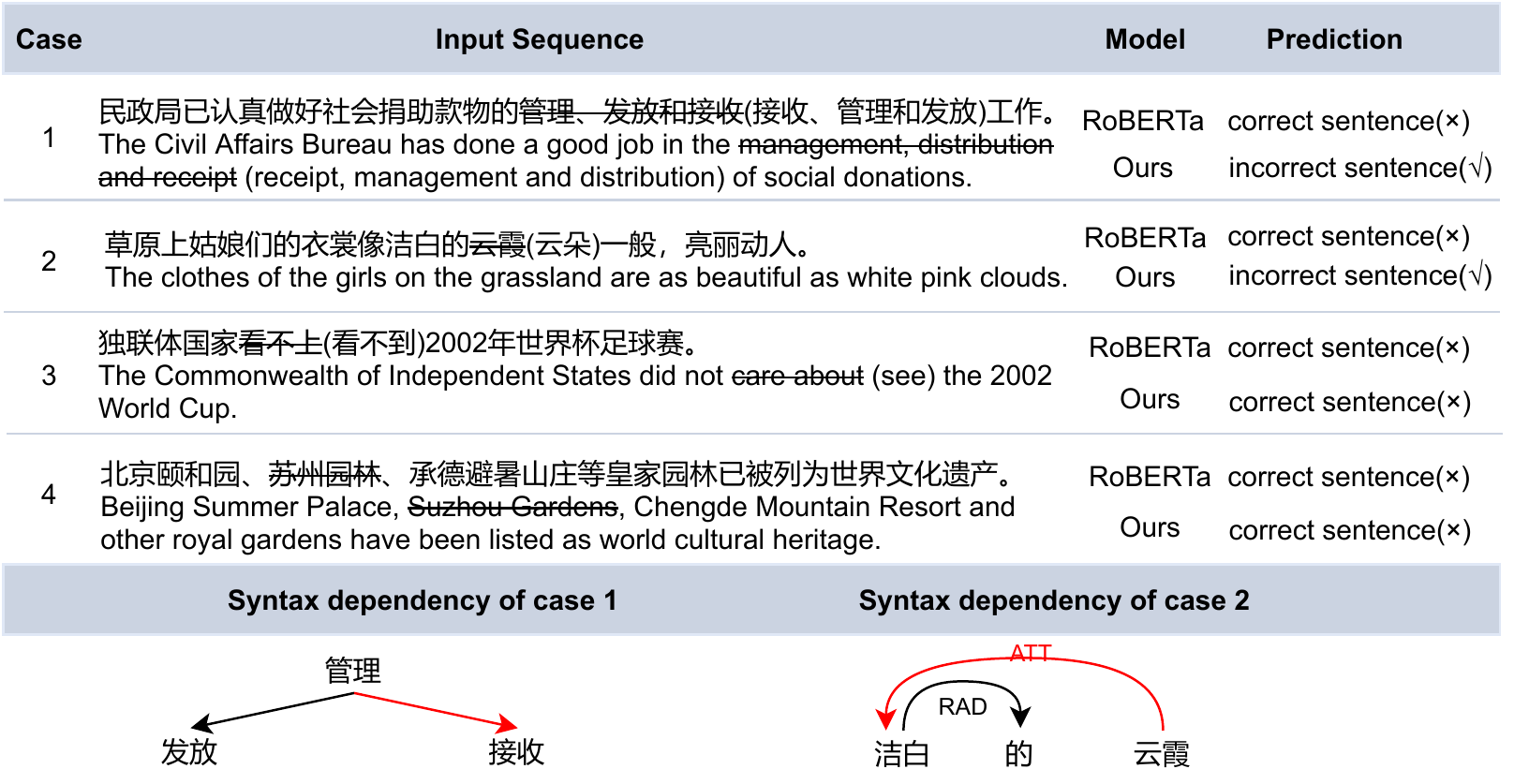} 
\caption{Case Study: we select several sentences with semantic errors, of which our model recognizes the first two, and the latter two are not. However, RoBERTa can not recognize all of them. The incorrect semantic in syntax dependency is marked as red.}
\label{case_study}
\end{figure*}

\subsection{Case Stusy}
We conduct a case study to demonstrate the effectiveness of learning syntactic information in the pre-training stage.
As shown in Figure \ref{case_study}, cases 1-2 are examples where our models give the correct answer, but RoBERTa does not. Cases 3-4 are examples where neither our models nor RoBERTa give the correct answer. From the syntax dependency of cases 1-2, we can see that our models potentially learn some dependencies during the pre-training stage. However, our model is also powerless for some examples that have little to do with syntactic dependencies. For example in case 3, \begin{CJK*}{UTF8}{gbsn}``看不上''\end{CJK*} has two meanings in Chinese: (1) \begin{CJK*}{UTF8}{gbsn}``瞧不起''\end{CJK*} (do not care) and (2) \begin{CJK*}{UTF8}{gbsn}``
看不见''\end{CJK*} (do not see). In case4, ``Suzhou gardens'' should be removed because Suzhou gardens are not royal gardens. The above errors require external knowledge to solve, and there is no way to rely solely on syntactic dependencies.

\section{Conclusion}
Unlike CGED and CSC, CSER is more difficult for humans to identify semantic errors and more dependent on syntactic information due to the complexity and variety of incorrect semantic sentences. This paper proposes the pre-trained model with several novel pre-training tasks that better learn syntactic dependency. Due to the vacancy of the dataset for CSER, we provide a high-quality dataset named CoCLSA. We fine-tune the pre-trained models on CoCLSA. The experimental results show that our models exceed existing pre-trained models and syntax-infused models. Our work could bring machines closer to the human level for the CSER task, and there is much room for improvement in the future.

\normalem
\newpage
\bibliography{anthology,custom}
\bibliographystyle{acl_natbib}

\appendix
\clearpage
\noindent\textbf{Appendix}
\section{Humans vs. Models}

\begin{table}[t]
\centering
\scalebox{0.9}{
\begin{tabular}{c|cccc}
\toprule
Model   & $P$& $R$& $F_1$ &$ACC$ \\ \hline
RoBERTa &73.7	&68.0 &70.7	&71.8\\
RoBERTa+DP(ours) & \textbf{77.5}	&70.0 &73.6 &74.8\\
RoBERTa+DP$^{\dagger}$(ours) & 75.3	&75.0 &75.1 &\textbf{75.0}\\
Human & 71.6 &\textbf{82.5} &\textbf{76.5}	&74.4 \\\bottomrule
\end{tabular}}
\caption{Human vs. Machine results on a small dataset} \label{histogram_table}
\end{table}
We sample 200 pieces of data from the test set in Table \ref{CoCLSA}, including 100 correct sentences and 100 incorrect sentences. We have plotted histograms of recall scores for different types of semantic errors in Figure \ref{histogram}. In this section, we calculate the precision, recall, F1, and accuracy score of humans and models for entire incorrect sentences shown in Table \ref{histogram_table}. 

Overall, models have higher precision scores and lower recall scores than humans. This may be because students have to choose a correct sentence from a multiple-choice question containing four options in the Chinese college entrance examination. This mindset leads people to think that most sentences contain semantic errors. Hence, humans have higher recall scores compared with models. It is worth noting that our method is comparable to humans in terms of accuracy score. However, RoBERTa is still far from the human level.

\section{Data Overlap}\label{overlap}
Data overlap means that the data in the training set and the test set are the same or highly similar. In this paper, we use the Levenshtein ratio as the similarity score between texts. We clean the data of the train set with a similarity score greater than 70\% between the train and test set. Because we find that sentences with a similarity score lower than 70\% can be considered to have no data overlap problem. We enumerate the top-5 sentence pairs with the highest similarity between cleaned train and test sets as shown in Table \ref{data_overlap}. In Case 1-2, the data in the train set and the test set are not similar. In Case 3-5, the sentence labels of the train set and the sentence labels of the test set are even different. 
\begin{table*}[t]
\renewcommand\arraystretch{1.3}
\scalebox{0.95}{
\begin{tabular}{llll}
\toprule
Case              & dataset & Sentence & Tag \\ \hline
\multirow{2}{*}{1} & train  & \begin{CJK*}{UTF8}{gbsn}在激烈的市场竞争中，博兰公司所缺乏的，一是创意不佳，二是资金不足。\end{CJK*}& $\times$   \\
& test&\begin{CJK*}{UTF8}{gbsn}在激烈的市场竞争中，很多企业所缺乏的，一是勇气不足，二是谋略不当。\end{CJK*}& $\times$ \\ \hline
\multirow{4}{*}{2} & \multirow{2}{*}{train}  & \begin{CJK*}{UTF8}{gbsn}互联网不仅能浏览信息、收发电子邮件，还可以提供网上视频点播和远程\end{CJK*}& \multirow{2}{*}{$\times$}   \\
&&\begin{CJK*}{UTF8}{gbsn}教学等智能化、个性化。\end{CJK*}& \\
& \multirow{2}{*}{test}&\begin{CJK*}{UTF8}{gbsn}宽带网络作为信息社会的主要纽带，它不仅能浏览信息，还可以提供网上\end{CJK*}& \multirow{2}{*}{$\times$} \\
&&\begin{CJK*}{UTF8}{gbsn}视频点播和远程教育等智能化、个性化。\end{CJK*}& \\\hline
\multirow{2}{*}{3} & train  & \begin{CJK*}{UTF8}{gbsn}劳动工资的改革，对某些吃惯“大锅饭”的职工，的确会感到不适应。
\end{CJK*}& $\times$   \\
& test&\begin{CJK*}{UTF8}{gbsn}某些吃惯“大锅饭”的职工对劳动工资制度的改革，的确会感到不适应。
\end{CJK*}& $\checkmark$ \\\hline
\multirow{4}{*}{4} & \multirow{2}{*}{train}  & \begin{CJK*}{UTF8}{gbsn}只有充分地对于一个问题的两方面的事实和论点加以叙述和比较，才能得\end{CJK*}& \multirow{2}{*}{$\times$}   \\
&&\begin{CJK*}{UTF8}{gbsn}到良好的结果，但这里不可能这样做。\end{CJK*}& \\
& \multirow{2}{*}{test}&\begin{CJK*}{UTF8}{gbsn}我们只有对一个问题的两方面的事实和论点加以充分地比较和叙述,才能得\end{CJK*}& \multirow{2}{*}{$\checkmark$} \\
&&\begin{CJK*}{UTF8}{gbsn}到良好的结果。\end{CJK*}& \\\hline
\multirow{4}{*}{5} & \multirow{2}{*}{train}  & \begin{CJK*}{UTF8}{gbsn}随着求职竞争的加剧，招聘企业不仅注重学历、文凭等硬指标，也日益看\end{CJK*}& \multirow{2}{*}{$\checkmark$}   \\
&&\begin{CJK*}{UTF8}{gbsn}重求职者的工作热情、责任心与沟通能力等“软指标”。\end{CJK*}& \\
& \multirow{2}{*}{test}&\begin{CJK*}{UTF8}{gbsn}随着竞争的加剧，招聘企业不仅注重求职者的工作热情、责任心与沟通能\end{CJK*}& \multirow{2}{*}{$\times$} \\
&&\begin{CJK*}{UTF8}{gbsn}力等“软指标”，也日益看重求职者的学历、文凭等硬指标。\end{CJK*}&  \\\bottomrule
\end{tabular}}
\caption{Top-5 sentence pairs with the highest similarity between train and test sets.} \label{data_overlap}
\end{table*}

\section{Text Error Detection} \label{text_error}
Text errors mainly include spelling errors, grammatical errors, and semantic errors. We enumerate some sentences including various error types for various tasks as shown in Table \ref{all_dataset}. Firstly, spelling errors are the easiest, mainly made by non-native speakers.
Grammatical errors are more complex than spelling errors, focusing on word order, redundant words, missing words, and bad selection. However, these errors often make the entire statement unintelligible and easy to identify. For example, the word order problem only focuses on the order of two random words, making the whole sentence unintelligible, and humans can easily recognize it. In addition, word order problems may also arise between two phrases or between two short sentences. Our semantic errors extend grammatical errors and are more complex by comparison. These semantic errors are often examined in Chinese examinations of junior and senior high schools. We give the following explanation of the reasons for various semantic errors:
\begin{itemize}
    \item Word Order: The right order is \begin{CJK*}{UTF8}{gbsn}``交流和融合''\end{CJK*} (exchange and fusion) due to the time sequence.
    \item Collocation: There is an inappropriate collocation between \begin{CJK*}{UTF8}{gbsn}``提高''\end{CJK*} (improve) and \begin{CJK*}{UTF8}{gbsn}``规模''\end{CJK*} (scale). The correct collocation is \begin{CJK*}{UTF8}{gbsn}``扩大规模''\end{CJK*} (expand the scale).
    \item Missing: The predicate verb \begin{CJK*}{UTF8}{gbsn}``具有''\end{CJK*} (has) has no corresponding object. The correct should be added the object \begin{CJK*}{UTF8}{gbsn}``的特点''\end{CJK*} (the feature of).
    \item Redundant: \begin{CJK*}{UTF8}{gbsn}``质疑''\end{CJK*} (raise questions) contains the meaning of \begin{CJK*}{UTF8}{gbsn}``提出''\end{CJK*} (raise).
    \item Confusion: This sentence mixes the two complete sentences together. The one is that \begin{CJK*}{UTF8}{gbsn}``由于资金不足''\end{CJK*} (Due to insufficient funds). The other is that \begin{CJK*}{UTF8}{gbsn}``由于资金的限制''\end{CJK*} (Due to the limit of funds).
    \item Fuzziness: \begin{CJK*}{UTF8}{gbsn}``晚上来''\end{CJK*} can be interpreted as \begin{CJK*}{UTF8}{gbsn}``晚上/来''\end{CJK*} (at night) or \begin{CJK*}{UTF8}{gbsn}``晚/上来''\end{CJK*} (late).
    \item Illogic: There is no causal relationship between the preceding and following sentences.
\end{itemize}

\begin{table*}[t]
\centering
\renewcommand\arraystretch{1.3}
\scalebox{0.9}{
\begin{tabular}{lll}
\toprule
Task & Error Type &Sentence \\ \hline
\multirow{2}{*}{CSC}  & \multirow{2}{*}{Spelling Errors} & \begin{CJK*}{UTF8}{gbsn}他被\sout{眼睛}(眼镜)蛇咬了。\end{CJK*}\\
& & He was bitten by a \sout{eye snake} (cobra). \\ \hline
\multirow{8}{*}{CGED} & \multirow{2}{*}{Word Order} & \begin{CJK*}{UTF8}{gbsn}没有解决这个问题，\sout{不能}人类(不能)实现更美好的将来。\end{CJK*}\\ 
&& Without addressing this problem, \sout{cannot} humanity (cannot) achieve a better future.\\\cline{2-3}
&\multirow{2}{*}{Redundant}&\begin{CJK*}{UTF8}{gbsn}\sout{并}这些意见差距固然不容易解决，但并不是不能解决的。\end{CJK*}\\ 
& &\sout{And} While these gaps in opinion are not easy to resolve, they are not insurmountable\\\cline{2-3}
&\multirow{2}{*}{Missing}&\begin{CJK*}{UTF8}{gbsn}随着经济的发展人们的想法也(在)改变。\end{CJK*}\\ & &As the economy develops, people's minds change.\\\cline{2-3}
&\multirow{2}{*}{Bad Selection} &\begin{CJK*}{UTF8}{gbsn}吸烟不但对自己的健康\sout{好处}(不好)，而且给非吸烟者带来不好的影响。\end{CJK*}\\ & &Smoking is not only \sout{good} (bad) for one's health, but also bad for non-smokers.\\\hline
\multirow{22}{*}{CSER} & \multirow{3}{*}{Word Order} & \begin{CJK*}{UTF8}{gbsn}任何一种文明的发展都是与其他文明碰撞、\sout{融合、交流}(交流、融合)的过程\end{CJK*}\\ 
&& The development of any civilization is a process of collision, \sout{fusion and exchange}\\
&&(exchange and fusion) with others. \\\cline{2-3}
& \multirow{3}{*}{Collocation} & \begin{CJK*}{UTF8}{gbsn}为了提高这次舞会的档次\sout{和规模}，举办方特邀中国人民解放军乐团现场演奏。\end{CJK*}\\ 
&&In order to improve the grade \sout{and scale} of this dance, the organizer specially invited\\
&&the Chinese People's Liberation Army Band to perform live. \\\cline{2-3}
& \multirow{3}{*}{Missing} & \begin{CJK*}{UTF8}{gbsn}这种治疗方法具有见效快，无副作用(的特点)，以达到标本兼治的目的。\end{CJK*}\\ 
&&This treatment method has (the feature of) quick effect and no side effects, so as to \\
&&achieve the purpose of treating both the symptoms and the root causes. \\\cline{2-3}
& \multirow{2}{*}{Redundant} & \begin{CJK*}{UTF8}{gbsn}有一部分网友却对雷锋及雷锋精神提出了各种各样的所谓\sout{质疑}(疑问)\end{CJK*}\\ 
&&Some netizens have raised various so-called questions about Lei Feng and his spirit.\\\cline{2-3}
& \multirow{3}{*}{Confusion} & \begin{CJK*}{UTF8}{gbsn}由于资金不足\sout{的限制}，学校计划修建的图书楼和医疗室只好暂缓施工。\end{CJK*}\\ 
&&Due to \sout{limit of} insufficient funds, the school's planned library building and medical \\
&&room had to be postponed.\\\cline{2-3}
& \multirow{2}{*}{Fuzziness} & \begin{CJK*}{UTF8}{gbsn}山上的水宝贵，我们把它留给\sout{晚上来}的人喝。\end{CJK*}\\ 
&&The water is precious, we leave it to people who come to drink \sout{at night} (late). \\\cline{2-3}
& \multirow{4}{*}{Illogic} & \begin{CJK*}{UTF8}{gbsn}如今的手机已不再是单纯的通信工具，\sout{因而}成为人们生活中的贴身伴侣，用\end{CJK*}\\ 
&&\begin{CJK*}{UTF8}{gbsn}来尽情表现个人品位。\end{CJK*}\\
&&Today's mobile phone is no longer a simple communication tool, \sout{so} it has become \\
&&a personal companion in people's life, used to express personal taste to the fullest.\\
\bottomrule
\end{tabular}}
\caption{Examples of different tasks. We cannot translate the incorrect semantic sentences in the CGED task, so we translate them into the correct semantics.} 
\label{all_dataset}
\end{table*}

\end{document}